\documentclass[runningheads]{llncs}

\usepackage[T1]{fontenc}
\usepackage{newtxtext,newtxmath} 
\usepackage{amsmath}
\usepackage{graphicx}
\usepackage{booktabs}
\usepackage{enumitem}
\usepackage{multirow}
\usepackage{bm}
\usepackage{bbm}
\usepackage{lineno}
\usepackage[dvipsnames]{xcolor}
\usepackage[table]{xcolor} 
\usepackage{colortbl}
\usepackage{enumitem}
\usepackage{pifont}
\usepackage[table]{xcolor}
\usepackage{float}

\usepackage{nicematrix}
\usepackage{hhline}
\usepackage{wrapfig}
\usepackage{booktabs} 
\usepackage{lipsum}   
\setlist[itemize]{label=\textbullet}
\definecolor{lightroyalblue}{RGB}{200,220,255}

\usepackage[normalem]{ulem}
\usepackage{array}        
\usepackage{makecell}     
\usepackage{microtype}    

\newcolumntype{C}[1]{>{\centering\arraybackslash}p{#1}}
\usepackage{appendix}

\usepackage{xcolor}
\usepackage{authblk}

\usepackage[colorlinks=true, citecolor=purple]{hyperref}
\definecolor{lavender}{RGB}{230,230,250}
\definecolor{lightblue}{RGB}{173, 216, 230}
\definecolor{tablegray}{gray}{0.92}

\usepackage{cleveref}
\begin{document}
\title{\textsc{PlugSI}: Plug-and-Play Test-Time Graph Adaptation for Spatial Interpolation}
\author{
Xuhang Wu\inst{1}$^\dag$\and
Zhuoxuan Liang\inst{1}$^\dag$ \and
Wei Li\inst{1}\thanks{Corresponding author; $^\dag$ Equal contribution.\\
This work was supported by Natural Science Foundation of Heilongjiang Province, grant number LH2023F020, and Supporting Fund of Intelligent Internet of Things and Crowd Computing, grant number B25029.} \and
Xiaohua Jia\inst{2} \and
Sumi Helal\inst{3}
}

\institute{
College of Computer Science and Technology, Harbin Engineering University, China \\
\email{\{xh.wu, zz.liang, wei.li\}@hrbeu.edu.cn}
\and
Department of Computer Science, City University of Hong Kong, China \\
\email{csjia@cityu.edu.hk}
\and
Department of Computer Science and Engineering, University of Bologna, Italy \\
\email{sumi.helal@unibo.it}
}

\maketitle              
%
\begin{abstract}
With the rapid advancement of IoT and edge computing, sensor networks have become indispensable, driving the need for large-scale sensor deployment. However, the high deployment cost hinders their scalability. To tackle the issues, Spatial Interpolation (SI) introduces virtual sensors to infer readings from observed sensors, leveraging graph structure. However, current graph-based SI methods rely on pre-trained models, lack adaptation to larger and unseen graphs at test-time, and overlook test data utilization.
To address these issues, we propose \textsc{PlugSI}, a plug-and-play framework that refines test-time graph through two key innovations.
First, we design an Unknown Topology Adapter (UTA) that adapts to the new graph structure of each small-batch at test-time, enhancing the generalization of SI pre-trained models.
Second, we introduce a Temporal Balance Adapter (TBA) that maintains a stable historical consensus to guide UTA adaptation and prevent drifting caused by noise in the current batch.
Empirically, extensive experiments demonstrate \textsc{PlugSI} can be seamlessly integrated into existing graph-based SI methods and provide significant improvement (e.g., a 10.81\% reduction in MAE).
\end{abstract}
\section{Introduction}
Ongoing IoT and edge computing progress has led sensor networks to play an essential role in fields including traffic flow~\cite{zhou2021informer}, air quality~\cite{belavadi2020air}, and solar energy~\cite{liu2022scinet}.
As a result, large-scale sensor deployment has emerged as a shared objective in diverse domains~\cite{liang2025darkfarseer}.
Nevertheless, the significant expenses involved in installing sensors pose a major obstacle in this context.
To alleviate this issue, Spatial Interpolation (SI), also known as Spatio-temporal Kriging~\cite{oliver1990kriging}, has emerged. SI introduces the concept of \textit{virtual sensors} to simulate those that would otherwise need to be deployed. Based on this concept, SI aims to infer the readings of virtual sensors from observed sensor data, thereby significantly reducing labor costs.

Abstracting sensor networks as graph structures is an effective approach, thus most previous deep learning-based methods~\cite{wu2021inductive,wu2021spatial,zheng2023increase,hu2023decoupling,li2024non,xu2025kits} are graph-based.
However, in the real world, sensor networks often expand continuously over time, causing the graphs to grow accordingly. This necessitates a consideration in SI models' design: \textit{the ability to generalize to larger graphs during test-time using only a single pre-training}. To this end, prior works~\cite{wu2021inductive,wu2021spatial,zheng2023increase,hu2023decoupling,li2024non,xu2025kits} have focused on pre-training: learning highly generalizable pre-trained models. 
However, all of these methods just focus on \textit{learning better representations during the pre-training phase}, while \textit{\textbf{the utilization of streaming graph structure data presents during test-time has been overlooked}}, which hinders SI model's performance.

Nevertheless, fully utilizing graph structure during testing is non-trivial. On the one hand, the data is streaming, meaning that it arrives in small batches. On the other hand, the graph structure is not fixed—the relationships between virtual and observed sensors may evolve over time with the streaming data. (e.g., as for traffic networks, a congestion event at one moment can change the correlations between nearby sensors in subsequent periods). In light of these considerations, we propose two \textbf{challenges}:
\begin{wrapfigure}{r}{0.45\textwidth} 
    \vspace{-0.5cm}
    \includegraphics[width=0.45\textwidth]{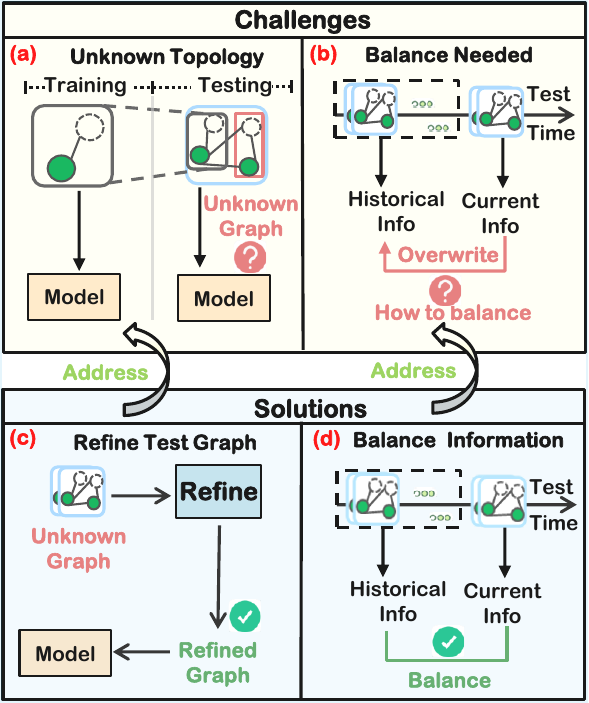} 
    \caption{Overall insight of our study. Fig.~1(a) depicts the unknown graph topology at test-time, 
and Fig.~2(c) shows it should be refined before inference. Meanwhile, Fig. 1(b) highlights current batch may overwrite historical refinement information. So Fig. 1(d) thinks the information of historical and current information should be balanced.
}
    \label{fig:introduce}
    \vspace{-0.5cm}
\end{wrapfigure}
\noindent \ding{182} \textbf{Unknown Graph-Topology Adaptation (\textit{Spatial Perspective}).}
During test-time, the inclusion of virtual sensors introduces new connections that alter and complicate the graph topology (see Fig. \ref{fig:introduce}(a)).
Moreover, as the streaming data continuously arrives, the spatial relationships between virtual and observed sensors dynamically evolve rather than remain fixed.
Consequently, the pre-trained model may struggle to handle these unseen and evolving spatial relationships, resulting in performance degradation as their topology uncertainty.
However, these new and evolving connections may capture meaningful spatial relationships between virtual and observed sensors, which are crucial for inferring virtual sensor readings.
Hence, designing a strategy that can effectively refine to these topological and relational dynamics during test is vital for SI (see Fig. \ref{fig:introduce}(c)).

\noindent \ding{183} \textbf{Balancing Historical and Current Information (\textit{Temporal Perspective}).} 
During test-time, data arrives sequentially as an stream of small-batches
~\cite{niu2022efficient,niu2023towards}. 
This streaming nature introduces an adaptation challenge: the adaptation must quickly respond to the current batch while preserving stability across the evolving data stream. In other words,
if overfitting to a small-batch with transient fluctuations (e.g., a sudden traffic anomaly caused by an accident) may overwrite the learned historical knowledge from previous batches,
it will impair the performance.
(see Fig. \ref{fig:introduce}(b)). 
Therefore, designing an adaptation method that achieves a balance between current responsiveness and historical consistency is a core focus of this study. (see Fig. \ref{fig:introduce}(d)).

Considering these challenges, fine-tuning models during deployment is often impractical—especially under streaming data~\cite{read2018concept,wares2019data}. In this study, we propose \textsc{PlugSI}, a plug-and-play, data-centric framework that externally augments pre-trained models by refining test-time graphs using a self-supervised task, without requiring access to training data or altering the pre-trained model. Specifically, \textbf{to address the first challenge}, we introduce the Unknown Topology Adaper (UTA). This design aims to enhance robustness and stability against unknown and changing topologies during testing. Motivated by the observation that unknown graph topology mainly arise from virtual nodes~\cite{liang2025darkfarseer}, UTA focuses on adapting virtual nodes using a Virtual Uncertainty Scorer (VUS) and an Adjacency Refiner (AR) that refines the edges.
This design achieves efficient, batch-wise adaptation for each test batch.
\textbf{To address the second challenge}, we propose a Temporal Balance Adapter (TBA), a lightweight online memory module. Unlike previous methods that preserve the pre-trained model parameters~\cite{tarvainen2017mean,xie2020unsupervised,sohn2020fixmatch}, TBA stores the adaptation information produced by UTA over several small-batches and gradually blends them with the current batch’s information. 
By balancing current adaptation and historical consistency, TBA ensures continuous yet stable test-time adaptation for streaming data and guides UTA in adapting the current batch.

We highlight the contributions of this paper as:
\begin{itemize}
    \item \textbf{New Perspective for SI:} We revisit SI from a test-time perspective. To this end, 
    we propose a plug-and-play framework \textsc{PlugSI}, which enhances the pre-trained SI models by leveraging test-time graph data and can be seamlessly plugged into existing graph-based SI methods. To our best knowledge, this work is the first to investigate the test-time design on SI.
    \item  \textbf{Innovative Solution:}, 
    We develop two complementary components: a UTA that adapts to unseen and dynamically changing graph structures within each test batch, and a TBA that ensures stable refinement by balancing current updates with accumulated historical information.
    \item \textbf{Comprehensive Experiments:}
    Experiments on four datasets spanning diverse domains demonstrate that \textsc{PlugSI} can consistently improve the existing graph-based SI models. Additionally, the experiments show that \textsc{PlugSI} is robust, scalable, and lightweight, indicating highly practical values.
\end{itemize}

The remainder of this paper is organized as follows: Section \ref{section_2} introduces the design details of \textsc{PlugSI} and provides complexity analysis (Section \ref{section_2_6}). Section \ref{section_3} presents the experimental design and results. Section \ref{section_4} discusses the limitations of \textsc{PlugSI} and our future directions. Section \ref{section_5} reviews related work. Finally, Section \ref{section_6} concludes the study.

\section{Methodology}\label{section_2}
\subsection{Preliminaries.}
In this work, we focus on enhancing the \textit{test-time} of SI by developing a plug-in module that enables the SI pre-trained model $\mathcal{F}$ to adapt test graph. The definition of SI~\cite{xu2025kits} please refer to Appendix \ref{section_Appdx}.

\noindent\textbf{Test-Time Graph Adaptation (TTGA).}
Let $\mathcal{B} = \{ \mathbf{B}_t \}_{t=1}^{T}$ denote a sequence of small-batches arriving over time.
Each small-batch at time $t$ is defined as $\mathbf{B}_t = (\mathbf{X}_t, \mathbf{A}_t)$,
where $\mathbf{X}_t \in \mathbb{R}^{B \times N \times P}$ represents readings from $N$ sensors over a temporal window of length $P$, with $B$ the batch size,
and $\mathbf{A}_t \in \mathbb{R}^{N \times N}$ denotes the adjacency matrix represents relationships among the $N$ sensors. 
The $N$ sensors include $N_o$ observed and $N_u$ virtual ones, where the readings of virtual sensors $\mathbf{X}^u_t$ are entirely missing.
Following the standard TTA protocol~\cite{li2023robustness}, the goal of TTGA is to perform online adaptation under streaming test data.
We learn an adaptation function $\mathcal{G}_{\mu}$, parameterized by $\mu$, to refine $\mathbf{A}_t$ as 
$\hat{\mathbf{A}}_t = \mathcal{G}_{\mu}(\mathbf{A}_t)$.
The $\mu$ are optimized by minimizing an Self-supervised task $\mathcal{L}$ over $\mathbf{B}_t$:
\begin{equation}
\label{eq:ttga_objective}
\mu^* = \arg\min_{\mu} \frac{1}{B} \sum_{b=1}^{B} \mathcal{L}(\mathcal{F}(\mathbf{X}_t^{u(b)}, \mathcal{G}_{\mu}(\mathbf{A}_t))).
\end{equation}
The optimized $\mathcal{G}_{\mu^*}$ is then applied to adapt $\mathbf{A}_t$ for inference $\mathbf{X}^u_t$.

\begin{figure}[htbp]
    \centering
    \includegraphics[width=1\textwidth]{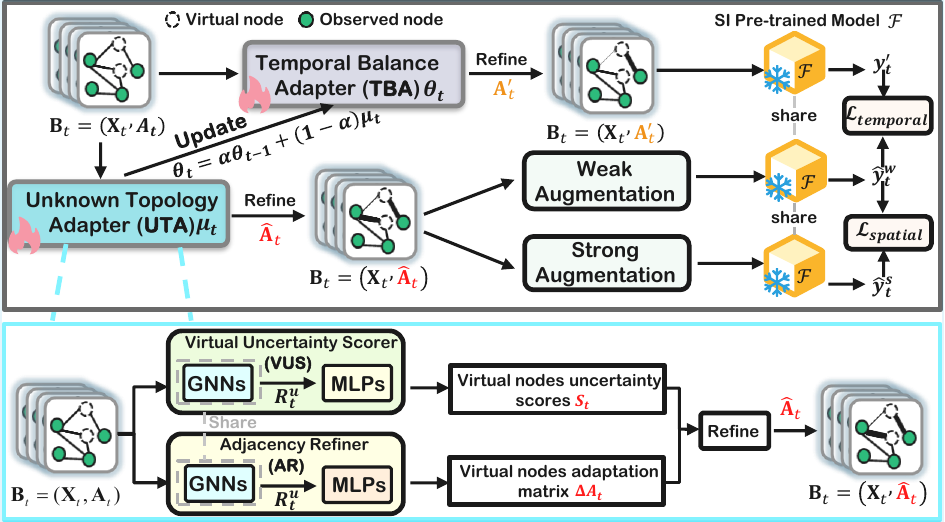}
    \caption{The overall framework of \textsc{PlugSI}. At a high level, it refines the local topology of virtual sensors in each small-batch $\mathbf{B}_t$, yielding a refined adjacency matrix $\mathbf{\hat{A}}_t$. \textsc{PlugSI} comprises two components: UTA, where VUS estimates topological uncertainty $\mathbf{S}_t$ for virtual sensors and AR derives the adaptation $\Delta\mathbf{A}_t$ to form $\mathbf{\hat{A}}_t$; and TBA, which integrates the current adaptation with accumulated historical information to generate updated guidance $\mathbf{A}'_t$ for the current batch and ensure stable online adaptation.
 }
    \label{fig:framework}
\end{figure}
\subsection{PLUGSI Overview}
The overall architecture of \textsc{PlugSI} is illustrated in Fig.~\ref{fig:framework}.
At a high level, \textsc{PlugSI} dynamically refines virtual sensors’ local topology for each small-batch in streaming data. Concretely, it consists of two interactive components: Unknown Topology Adapter (UTA) and Temporal Balance Adapter (TBA). 
Within the UTA module, VUS estimates topological uncertainty scores for virtual sensors, which are then used by AR to refine the adjacency matrix for each small-batch.
TBA uses accumulated historical information to guide UTA’s adaptation of the current batch, combining it with the current small-batch update to ensure stable online adaptation.

\textbf{UTA.}
~UTA refines the connections of virtual nodes in an effective-yet-efficient manner, and consists of two modules: the Virtual Uncertainty Scorer (VUS) and the Adjacency Refiner (AR). Firstly, VUS extracts contextual spatio-temporal representations $R_t^u$ and generates topological uncertainty scores $S_t$ for virtual nodes, highlighting which nodes need prioritized adaptation. Subsequently, AR uses $R_t^u$ with the $S_t$ to refine the adjacency matrix, adapting virtual nodes to the spatial relationships of the current small-batch. This design enables efficient, batch-wise adaptation, refining the test data without modifying the pre-trained model. See Section~\ref{sec:UTA} for the details.

\textbf{TBA.}
~TBA is a lightweight memory module designed to balance immediate adaptation with the preservation of historical information. First, it stores the adaptation information accumulated from UTA across multiple small-batches, serving as a stable reference for past information. Next, TBA integrates the short-term adaptation from the current small-batch, reconciling new updates with historical information. Further, it updates its parameters in a weighted manner to smoothly combine past and present information. This design ensures consistent, batch-wise adaptation to streaming test data while retaining long-term stability. See Section~\ref{sec:TBA} for the details.

\subsection{Unknown Topology Adapter (UTA)}
\label{sec:UTA}
To address \textbf{Unknown Graph-Topology Adaptation} (the first challenge), we propose \textbf{UTA} in this section, which dynamically adapts to unseen and evolving test graphs in each small-batch.

Generalizing to larger and unknown graphs at test-time represents a core challenge in SI. To enhance robustness against such topological changes, the Increment Training Strategy~\cite{xu2025kits} has been proposed. Unlike prior works that simulate virtual nodes by masking observed ones during training~\cite{wu2021inductive,zheng2023increase}, this strategy expands the graph by inserting virtual nodes, thereby better emulating the potential topological variations encountered during test-time.
However, it incurs high computational cost and struggles to generalize to unseen graphs.
In the graph domain, this challenge is known as  \textit{structure shifts}~\cite{zhu2021shift,liu2024pairwise}. Test-Time Adaptation (TTA)~\cite{bao2024matcha,zhang2024fully} has emerged as a promising approach, where most methods fine-tune a pre-trained model during test-time. However, such methods are not always practical in real-world deployment when confronted with a stream of online test samples.

\noindent\textbf{Virtual Uncertainty Scorer (VUS).} Inspired by~\cite{jin2022empowering}, which refines a test graph by injecting perturbations into the topology across all nodes. But in SI, we argue that observed nodes constitute the factual foundation for inference, whereas structure shifts arise from the virtual nodes and their induced new connections, inspired by~\cite{liang2025darkfarseer}. Accordingly, our method adopts this core idea only through transformation applied to the virtual nodes. We first employ a simple GNN~\cite{kipf2016semi} to extract contextual spatio-temporal representations $\mathbf{R_t} \in \mathbb{R}^{B \times N \times D}$, where $D$ denotes the hidden state dimension. These representations $\textbf{R}_t$ are split into virtual $\mathbf{R}_t^u \in \mathbb{R}^{N_u \times D} $ and observed $\mathbf{R}_t^o \in \mathbb{R}^{B \times N_o \times D}$. We only focus on $N_u$ virtual nodes, then $\mathbf{R}_t^u \in \mathbb{R}^{B \times N_u \times D}$ are fed into a two-layer MLPs:
\begin{equation}
\label{eq:score}
\mathbf{S}_t = \sigma(\text{MLP}(\text{GNN}(\mathbf{X}_t, \mathbf{A}_t)[N_u])),
\end{equation}
where $\sigma(\cdot)$ is the Sigmoid activation function and $\mathbf{S}_t \in [0,1]^{N_u}$ serves to estimate the topological uncertainty introduced by virtual nodes, thereby enabling a prioritized adaptation of the test graph.
Building on adaptation score $\mathbf{S}_t$ obtained in
Eq. \eqref{eq:score}, we discuss how to use $\mathbf{S}_t$ in the
following section.

\noindent\textbf{Adjacency Refiner (AR).} 
Building upon the score $\textbf{S}_t$, we further address the structure shifts by refining $\mathbf{A}_t$. 
A common strategy is to add or delete edges~\cite{zeng2025graph,li2022reliable}. However, such operations may distort the true relationships among nodes. 
Inspired by~\cite{jin2022empowering}, which adjusts the adjacency matrix by injecting learnable perturbations, we note that it neglects the weighted relationships among neighbors. 
To address this limitation, we propose AR. 
Specifically, AR leverages the contextual representations $\mathbf{R}_t^u$ obtained from the VUS to characterize the neighborhood context of each virtual node. 
Let $\mathcal{N}_v(u)$ denote the set of neighbors connected to the virtual node $u$. 
The $\mathbf{R}_t^u$ are then fed into a two-layer MLPs to produce an adaptation matrix $\Delta\mathbf{A}_t \in \mathbb{R}^{N \times N}$, where only entries corresponding to edges $(u, v)$ with $v \in \mathcal{N}_v(u)$ are updated. 
The refined adjacency matrix $\mathbf{\hat{A}}_t$ is computed as:
\begin{equation}
\label{eq:refine}
(\mathbf{\hat{A}}_t)_{uv} =
\begin{cases}
(\mathbf{A}_t)_{uv} + (\mathbf{S}_t)_u \cdot (\Delta\mathbf{A}_t)_{uv}, & \text{if } v \in \mathcal{N}_v(u), \\[4pt]
(\mathbf{A}_t)_{uv}, & \text{otherwise}.
\end{cases}
\end{equation}
In this way, the $\mathbf{\hat{A}}_t$ which adapts the connections of virtual nodes, is the adapted structure referenced in  Eq. (\ref{eq:ttga_objective}), 
which is then provided as input to the fixed pre-trained model $\mathcal{F}$, enabling more robust performance against the dynamic graph structures encountered in each small-batch.
\subsection{Temporal Balance Adapter (TBA)}
\label{sec:TBA}
To address \textbf{Balancing Historical and Current Information}  (the second challenge), we propose TBA in this section, which aims to capture multi-batch relationships and guide each small-batch’s adaptation.

In TTA, balancing the integration of newly information with the preservation of historical information remains a core challenge under online streaming data. A weight-averaged model~\cite{tarvainen2017mean,xie2020unsupervised,sohn2020fixmatch}, constructed in the same architecture as the pre-trained model, is employed to retain the historical information of pre-trained model and mitigate such issues.
Unlike these works, our approach maintains the adaptation information accumulated of UTA from multiple small-batches, rather than the pre-trained model. We conceptualize TBA as the weight-averaged model with parameters $\theta$, and UTA as the pre-trained model, adopting this core idea in a simple yet effective manner. Our goal is to update TBA to integrate short-term information from the current small-batch while maintaining long-term consistency with historical information. At time step $t$, $\mu_t$ denotes UTA's parameters adapted from the current batch $\mathbf{B}_t$, and $\theta_{t-1}$ represents the historical TBA's parameters. Mathematically, this balance can be formulated as:
\begin{equation} 
\label{eq:balance}
\theta_t = \arg\min_{\theta_t} \left( \alpha ||\theta_t - \theta_{t-1}||^2_2 + (1 - \alpha)||\theta_t - \mu_t||^2_2 \right),
\end{equation}
where $\alpha$ is an information balance coefficient. 
By minimizing  Eq.~\eqref{eq:balance}, we can obtain the parameter update for $\theta_t$:
\begin{equation}
\theta_t = \alpha \theta_{t-1} + (1-\alpha)\mu_t,
\label{eq:meanteacher}
\end{equation}
By the way, it allows TBA to smoothly incorporate new adaptation information while preserving stability from historical parameters.

\subsection{Optimization Objective}
As discussed earlier, the proposed framework \textsc{PlugSI} aims to improve the model generalization and robustness by refining the test graph, the primary obstacle for any data-centric graph transformation is the absence of ground-truth labels for virtual nodes. Self-supervised learning (SSL) techniques~\cite{sun2020test}, however, pave the way for providing self-supervision for our test-time graph adaptation. 

In this work, we propose a self-supervised task which assesses the temporal and spatial consistency of UTA. 
To be specific, for each refined graph $\hat{\mathbf{A}}_t$, we apply a feature-masking augmentation function $H(\cdot; r)$ with two different mask rates, $r_w$ and $r_s$ ($r_w < r_s$), to generate weak and strong augmented views. 
We then feed these two views into the fixed pre-trained model $\mathcal{F}$ to obtain the corresponding inferences:

\begin{equation}
\mathbf{\hat{y}}_t^{\{w,s\}} =
\mathcal{F}\!\big(H(\mathbf{X}_t^u; r_{\{w,s\}}, \hat{\mathbf{A}}_t), \hat{\mathbf{A}}_t\big),
\label{eq:prediction_consistency}
\end{equation}

\noindent where $\mathbf{\hat{y}}_i^{w}$ and $\mathbf{\hat{y}}_i^{s}$ denote the inferences from weak and strong augmentations, respectively. 
An effective adapter $\mathcal{G}_{\mu}$ should produce refined structures that are robust to different augmentation strengths, leading to stable inferences. 
Then, the learning objective of spatial consistency can be formulated as:

\begin{equation}
\mathcal{L}_{\text{spatial}}
=
\frac{1}{B} \sum_{b=1}^{B} 
\frac{1}{N_u \cdot P} 
\big\lVert 
\mathbf{\hat{y}}_t^{w(b)} - \mathbf{\hat{y}}_t^{s(b)}
\big\rVert_1
\label{eq:spatial_loss}
\end{equation}

To prevent the adaptation on the current batch $B_t$ from deviating from historical adaptation information, we apply TBA to obtain $\tilde{\mathbf{A}}_t$ and compute the temporal loss:
\begin{equation}
\mathcal{L}_{\mathrm{temporal}}
=
\frac{1}{B} \sum_{b=1}^{B} 
\frac{1}{N_u \cdot P} \big\lVert 
\mathcal{F}(\mathbf{X}_t^{u(b)}, \tilde{\mathbf{A}}_t) - \mathbf{\hat{y}}_t^{w(b)}
\big\rVert_1
\label{eq:temporal_loss}
\end{equation}

Furthermore, the learning objective of test-time can be written as:
\begin{equation}
\mathcal{L} = \mathcal{L}_{\text{spatial}} + \mathcal{L}_{\mathrm{temporal}} + \lambda \text{Reg}(\Delta \mathbf{A}_t)
\label{eq:loss}
\end{equation}
where $\lambda$ is a trade-off parameter and $\text{Reg}(\Delta\mathbf{A}_t)$ serves as a regularization term to ensure we do
not heavily violate the original graph structure.
\subsection{Complexity Analysis}\label{section_2_6}
\textbf{$\star$ Remarks.} In conclusion, the complexities grow \textit{linearly} with the graph size $N$ and the average degree of virtual nodes $d$, which confirms \textsc{PlugSI} is highly scalable.  
\subsubsection{Time Complexity.}
The time complexity of \textsc{PlugSI} is dominated by its UTA and TBA components.
In UTA, the VUS (Eq.~\eqref{eq:score}) applies an MLP to the $N_u$ virtual node representations, incurring $O(N_u)$ cost, while the AR module (Eq.~\eqref{eq:refine}) performs refinements only on edges connected to virtual nodes, resulting in $O(N_u d)$ complexity.
During batch optimization, the TBA update (Eq.~\eqref{eq:meanteacher}) and loss computation (Eq.~\eqref{eq:loss}) both involve operations proportional to the adapted edges, each with $O(N_u d)$ cost.
Overall, disregarding the parameters of the GNN, adapting a small-batch incurs a total time complexity of $O(B N_u d)$.

\subsubsection{Space Complexity.}
For \textsc{PlugSI}, a small-batch of sensor readings with shape $[B, N, P]$ requires $O(BNP)$ memory, while intermediate representations produced by the GNN consume $O(BND)$ space.
Additional memory is used to store sparse adaptation-related variables and TBA parameters, which scale with the number of adapted edges and require $O(BN_u d)$ space.
Thus, excluding the GNN parameters, the overall space complexity is ${O}(BNP + BND + BN_u d)$.

\section{Experiments and Results}\label{section_3}
We conduct comprehensive experiments to evaluate the effectiveness of the proposed \textsc{PlugSI} framework. 
Specifically, we aim to address the following research questions:
\begin{itemize}
    \item \textbf{RQ1:} How does \textsc{PlugSI} enhance the popular SI methods across multiple datasets?
    \item \textbf{RQ2:} How effective are the individual components of \textsc{PlugSI}?
    \item \textbf{RQ3:} How do key parameters affect \textsc{PlugSI}'s performance?
    \item \textbf{RQ4:} How robust is \textsc{PlugSI} under diverse virtual nodes scenarios, such as different proportion of virtual nodes and distribution patterns (e.g., random or regional)?
    \item \textbf{RQ5:} How efficient is \textsc{PlugSI} during inference compared to the baseline models, and what is the computational overhead introduced by the refinement process?
\end{itemize}

\begin{wraptable}{r}{0.5\linewidth}
\vspace{-1cm}
\centering
\caption{The overall information for datasets.}
\label{tab:selected_datasets}
\vspace{6pt}
\resizebox{0.5\textwidth}{!}{%
\begin{tabular}{c|c|c|c|c}
\toprule
\multirow{2}{*}{\textbf{Datasets}} & 
  \textbf{Traffic Speed} &
  \textbf{Traffic Flow} &
  \textbf{Air Quality} &
  \textbf{Solar Power} \\ \cmidrule(r){2-2} \cmidrule(r){3-3} \cmidrule(r){4-4} \cmidrule(l){5-5}
 & 
  \textbf{METR-LA} &
  \textbf{PEMS07} &
  \textbf{AQI-36} &
  \textbf{NREL-AL} \\ \midrule
\textbf{Regions} &
  Los Angeles &
  California &
  Beijing &
  Alabama \\
\textbf{Nodes} &
  207 &
  883 &
  36 &
  137 \\ \midrule 
\textbf{Timesteps} &
  34,272 &
  28,224 &
  8,759 &
  105,120 \\
\textbf{Intervals} &
  5min &
  5min &
  1hour &
  5min \\
\textbf{Start time} &
  3/1/2012 &
  5/1/2017 &
  5/1/2014 &
  1/1/2016 \\ \bottomrule
\end{tabular}%
}
\vspace{-1cm}
\end{wraptable}
\subsection{Experimental Settings}

\paragraph{\textbf{Datasets}.}
We employ 4 public datasets spanning four real-world application scenarios: traffic speed, traffic flow, air quality and solar power. Detailed dataset statistics are provided in Table \ref{tab:selected_datasets}. 
\paragraph{\textbf{Metrics}.}
We mainly adopt the Mean Absolute Error (MAE), Mean Relative Error (MRE), and Mean Absolute Percentage Error (MAPE) as evaluation metrics~\cite{cao2018brits,cini2021filling}. Among them, MAE and MRE evaluate relative deviations, while MAPE reflects absolute deviations. A MAPE below 100\% indicates that the model’s average prediction error is smaller than the corresponding true value, confirming its \textit{precision and practical effectiveness}. Lower values indicate better performance.

\paragraph{\textbf{Baselines}.}
Our framework is evaluated on seven popular graph-based SI methods to assess its effectiveness. These baselines include: KCN~\cite{appleby2020kriging}, IGNNK~\cite{wu2021inductive}, SATCN~\cite{wu2021spatial}, INCREASE~\cite{zheng2023increase}, DualSTN~\cite{hu2023decoupling}, KCP~\cite{li2024non} and KITS~\cite{xu2025kits}.
\paragraph{\textbf{Implementation Details}.}
To evaluate \textsc{PlugSI} at test-time, we follow the standard training process~\cite{wu2021inductive} to obtain a fixed pre-trained model.
And following the standard TTA protocol~\cite{li2023robustness}, we immediately perform inference upon receiving each small-batch. Concretely, for all datasets, we follow a fixed 7/2/1 split into training, validation, and test sets without shuffling, maintaining an observed-to-virtual nodes ratio of 1:1 following KITS~\cite{xu2025kits}. And we generate random masks to split all nodes into observed and virtual nodes.
For fairness, all baselines are evaluated with the same set of virtual nodes on each dataset.
Each experiment is repeated four times with a fixed random seed. We apply min–max normalization using statistics computed from the training set, which are then applied to the validation and test sets (avoiding data leakage). 
The time window $p$ is set to 24 and batch size is set to 32. 
Our code is implemented using Python 3.10.8 and PyTorch 2.2.1, and all experiments are conducted on a device equipped with dual NVIDIA GeForce RTX 4090 GPUs, a 12th Gen Intel(R) Core(TM) i9-12900K CPU, and 64GB of RAM, running Ubuntu 20.04 LTS.

\subsection{Comparison Results (Q1)}
\newcommand{\whitecell}[1]{\cellcolor{white}#1}
\begin{table*}
\centering
\caption{Performance over the test graphs on different backbone graph-based models.
All results are reported as mean $\pm$ standard deviation and \textbf{bold} numbers indicate improvements.
"–" denotes that the corresponding methods require GPS coordinates or complete distance information, which are unavailable in the PEMS07 dataset.}
\label{tb:allresult} 
\resizebox{\textwidth}{!}{%
\begin{tabular}{cccccccccccccccc} 
\toprule
\multirow{2}{*}{\textbf{Backbones}} & \multicolumn{3}{c}{\textbf{AQI-36}} & & \multicolumn{3}{c}{\textbf{METR-LA}} & & \multicolumn{3}{c}{\textbf{PEMS07}} & & \multicolumn{3}{c}{\textbf{NREL-AL}} \\ 
\cline{2-4}\cline{6-8}\cline{10-12}\cline{14-16}
 & \textbf{MAE} & \textbf{MRE} & \textbf{MAPE} & & \textbf{MAE} & \textbf{MRE} & \textbf{MAPE} & & \textbf{MAE} & \textbf{MRE} & \textbf{MAPE} & & \textbf{MAE} & \textbf{MRE} & \textbf{MAPE} \\ 
\midrule
KCN & 23.962{\footnotesize$\pm$0.284} & 0.349{\footnotesize$\pm$0.004} & 0.719{\footnotesize$\pm$0.006}& & 8.382{\footnotesize$\pm$0.054} & 0.159{\footnotesize$\pm$0.001} & 0.254{\footnotesize$\pm$0.001} & & \whitecell{-} & \cellcolor{white} - & \cellcolor{white}- & & 5.682{\footnotesize$\pm$0.043} & 0.436{\footnotesize$\pm$0.007} & 0.968{\footnotesize$\pm$0.003} \\
\rowcolor{lightroyalblue}\textsc{+ PlugSI} & \textbf{21.847{\footnotesize$\pm$0.252}} & \textbf{0.319{\footnotesize$\pm$0.003}}& \textbf{0.661{\footnotesize$\pm$0.007}}& & \textbf{7.962{\footnotesize$\pm$0.043}} & \textbf{0.152{\footnotesize$\pm$0.001}} & \textbf{0.239{\footnotesize$\pm$0.002}} & & \whitecell{-} & \whitecell{-} & \whitecell{-} & & \textbf{5.511{\footnotesize$\pm$0.039}} & \textbf{0.419{\footnotesize$\pm$0.006}} & \textbf{0.942{\footnotesize$\pm$0.007}} \\ 
\midrule
IGNNK & 22.652{\footnotesize$\pm$0.469}& 0.325{\footnotesize$\pm$0.005} & 0.796{\footnotesize$\pm$0.008} & & 6.412{\footnotesize$\pm$0.022} & 0.105{\footnotesize$\pm$0.001} & 0.175{\footnotesize$\pm$0.001} & & 80.457{\footnotesize$\pm$0.596} & 0.258{\footnotesize$\pm$0.024} & 0.972{\footnotesize$\pm$0.006} & & 5.613{\footnotesize$\pm$0.032} & 0.450{\footnotesize$\pm$0.009} & 0.970{\footnotesize$\pm$0.000} \\
\rowcolor{lightroyalblue}\textsc{+ PlugSI} & \textbf{20.840{\footnotesize$\pm$0.453}}& \textbf{0.303{\footnotesize$\pm$0.005}} & \textbf{0.747{\footnotesize$\pm$0.010}} & & \textbf{6.015{\footnotesize$\pm$0.018}} & \textbf{0.103{\footnotesize$\pm$0.001}} & \textbf{0.161{\footnotesize$\pm$0.002}}& & \textbf{76.434{\footnotesize$\pm$0.480}} & \textbf{0.248{\footnotesize$\pm$0.026}} & \textbf{0.928{\footnotesize$\pm$0.004}} & & \textbf{5.406{\footnotesize$\pm$0.036}} & \textbf{0.427{\footnotesize$\pm$0.011}} & \textbf{0.931{\footnotesize$\pm$0.001}} \\ 
\midrule
SATCN & 21.841{\footnotesize$\pm$0.940}& 0.539{\footnotesize$\pm$0.084} & 0.774{\footnotesize$\pm$0.013} & & 7.132{\footnotesize$\pm$0.033} & 0.116{\footnotesize$\pm$0.004} & 0.172{\footnotesize$\pm$0.001} & & 87.592{\footnotesize$\pm$0.647} & 0.305{\footnotesize$\pm$0.035} & 0.979{\footnotesize$\pm$0.002} & & 5.728{\footnotesize$\pm$0.040} & 0.453{\footnotesize$\pm$0.014} & 0.975{\footnotesize$\pm$0.003} \\
\rowcolor{lightroyalblue}\textsc{+ PlugSI} & 19.672{\footnotesize$\pm$0.857}& \textbf{0.501{\footnotesize$\pm$0.103}} & \textbf{0.713{\footnotesize$\pm$0.009}} & & \textbf{6.757{\footnotesize$\pm$0.035}} & \textbf{0.113{\footnotesize$\pm$0.002}} & \textbf{0.165{\footnotesize$\pm$0.001}} & & \textbf{72.268{\footnotesize$\pm$0.603}} & \textbf{0.238{\footnotesize$\pm$0.027}} & \textbf{0.928{\footnotesize$\pm$0.002}} & & \textbf{5.269{\footnotesize$\pm$0.032}} & \textbf{0.423{\footnotesize$\pm$0.010}} & \textbf{0.906{\footnotesize$\pm$0.003}} \\ 
\midrule
INCREASE & 22.983{\footnotesize$\pm$0.537}& 0.256{\footnotesize$\pm$0.013} & 0.872{\footnotesize$\pm$0.011} & & 7.804{\footnotesize$\pm$0.048} & 0.134{\footnotesize$\pm$0.002} & 0.244{\footnotesize$\pm$0.003} & & 91.753{\footnotesize$\pm$0.682} & 0.306{\footnotesize$\pm$0.031} & 0.804{\footnotesize$\pm$0.005} & & 5.579{\footnotesize$\pm$0.042} & 0.458{\footnotesize$\pm$0.006}& 0.974{\footnotesize$\pm$0.005} \\
\rowcolor{lightroyalblue}\textsc{+ PlugSI} & \textbf{21.842{\footnotesize$\pm$0.509}}& \textbf{0.235{\footnotesize$\pm$0.008}} & \textbf{0.799{\footnotesize$\pm$0.007} }& & \textbf{7.504{\footnotesize$\pm$0.042}} & \whitecell{0.142{\footnotesize$\pm$0.002}} & \textbf{0.240{\footnotesize$\pm$0.001}} & & \textbf{88.165{\footnotesize$\pm$0.652}} & \whitecell{0.311{\footnotesize$\pm$0.025}} & \textbf{0.810{\footnotesize$\pm$0.005} }& & \textbf{5.301{\footnotesize$\pm$0.037}} & \textbf{0.433{\footnotesize$\pm$0.004}} & \textbf{0.936{\footnotesize$\pm$0.001}} \\ 
\midrule
DualSTN & 22.536{\footnotesize$\pm$0.395}& 0.221{\footnotesize$\pm$0.011}& 0.411{\footnotesize$\pm$0.005} & & 8.441{\footnotesize$\pm$0.036} & 0.145{\footnotesize$\pm$0.020} & 0.219{\footnotesize$\pm$0.005} & & 90.862{\footnotesize$\pm$0.545} & 0.299{\footnotesize$\pm$0.031} & 0.853{\footnotesize$\pm$0.009} & & 4.502{\footnotesize$\pm$0.041}& 0.362{\footnotesize$\pm$0.014} & 0.947{\footnotesize$\pm$0.002} \\
\rowcolor{lightroyalblue}\textsc{+ PlugSI} & \textbf{21.232{\footnotesize$\pm$0.285}}& \textbf{0.219{\footnotesize$\pm$0.009}} & \textbf{0.391{\footnotesize$\pm$0.006}} & & \textbf{7.614{\footnotesize$\pm$0.041}}& \textbf{0.140{\footnotesize$\pm$0.018}} & \textbf{0.209{\footnotesize$\pm$0.007}}& & \textbf{83.427{\footnotesize$\pm$0.493}} & \textbf{0.281{\footnotesize$\pm$0.028}} & \textbf{0.790{\footnotesize$\pm$0.005}} & & \textbf{4.251{\footnotesize$\pm$0.039}} & \textbf{0.349{\footnotesize$\pm$0.009}} & \whitecell{0.951{\footnotesize$\pm$0.008}} \\ 
\midrule
KCP & 20.187{\footnotesize$\pm$0.359} & 0.259{\footnotesize$\pm$0.010} & 0.378{\footnotesize$\pm$0.008} & & 8.867{\footnotesize$\pm$0.044} & 0.138{\footnotesize$\pm$0.005} & 0.200{\footnotesize$\pm$0.008} & & 78.941{\footnotesize$\pm$0.527} & 0.261{\footnotesize$\pm$0.031} & 0.950{\footnotesize$\pm$0.003} & & 1.621{\footnotesize$\pm$0.025} & 0.259{\footnotesize$\pm$0.005} & 0.720{\footnotesize$\pm$0.000} \\
\rowcolor{lightroyalblue}\textsc{+ PlugSI} & \textbf{18.374{\footnotesize$\pm$0.314}} & \textbf{0.236{\footnotesize$\pm$0.008}} & \textbf{0.348{\footnotesize$\pm$0.010}} & & \textbf{8.494{\footnotesize$\pm$0.039}} & \textbf{0.125{\footnotesize$\pm$0.005}}& \textbf{0.184{\footnotesize$\pm$0.010}} & & \textbf{74.892{\footnotesize$\pm$0.522}} & \textbf{0.248{\footnotesize$\pm$0.027}} & \textbf{0.897{\footnotesize$\pm$0.006}} & & \textbf{1.556{\footnotesize$\pm$0.034}} & \textbf{0.256{\footnotesize$\pm$0.005}} & \whitecell{0.724{\footnotesize$\pm$0.003}} \\ 
\midrule
KITS & 22.339{\footnotesize$\pm$0.397}  & 0.295{\footnotesize$\pm$0.006} & 0.844{\footnotesize$\pm$0.003} & & 6.425{\footnotesize$\pm$0.020} & 0.105{\footnotesize$\pm$0.000} & 0.176{\footnotesize$\pm$0.002} & & 75.958{\footnotesize$\pm$0.401} & 0.249{\footnotesize$\pm$0.019} & 0.692{\footnotesize$\pm$0.006} & & 3.547{\footnotesize$\pm$0.026} & 0.337{\footnotesize$\pm$0.006} & 0.958{\footnotesize$\pm$0.002} \\
\rowcolor{lightroyalblue}\textsc{+ PlugSI} & \textbf{19.924{\footnotesize$\pm$0.204}} & \textbf{0.263{\footnotesize$\pm$0.004}}& \textbf{0.548{\footnotesize$\pm$0.006}} & & \textbf{6.021{\footnotesize$\pm$0.019}}& \textbf{0.104{\footnotesize$\pm$0.000}} & \textbf{0.162{\footnotesize$\pm$0.001}} & & \textbf{72.920{\footnotesize$\pm$0.395}} & \textbf{0.240{\footnotesize$\pm$0.022} }& \whitecell{0.705{\footnotesize$\pm$0.007}} & & \textbf{3.372{\footnotesize$\pm$0.021}} & \textbf{0.307{\footnotesize$\pm$0.009}} & \textbf{0.908{\footnotesize$\pm$0.006}} \\
\bottomrule
\end{tabular}
}
\end{table*}

To study \textbf{Q1}, we conduct comparative experiments over the test graphs using seven graph-based backbones across four real-world datasets from different domains, as shown in Table~\ref{tb:allresult}. Overall, \textsc{PlugSI} generally delivers great performance and verifies the outstanding effectiveness for refining graph data at test-time to serve better backbones adaptation ability.

Notably, for the datasets with extremely small (36 nodes) and extremely large (883 nodes) graph sizes, i.e., AQI-36 and PEMS07, our model consistently improves the test performance of all backbones, demonstrating its ability to effectively refine graph structures of varying scales. Meanwhile, on the long-horizon dataset NREL-AL (105,120 timesteps), our module consistently enhances the performance of most backbone models. This result highlights the capability of TBA to capture multiple small-batches dependencies and to stably balance historical and current information, validating its robustness under continuously streaming test data. Moreover, even for the state-of-the-art method KITS~\cite{xu2025kits}, which has been designed to closely simulate the potential graph structures at test-time, the refined graph still enhances its performance. This further demonstrates \textsc{PlugSI}'s superior ability to adapt to unknown graph topologies. On the traffic datasets METR-LA and PEMS07, \textsc{PlugSI} slightly reduces the inference ability of INCREASE. This may be attributed to our structural adaptation, which essentially optimizes spatial relations and may interfere with INCREASE’s capacity to learn other relational dependencies specific to traffic data.
\subsection{Ablation Study (Q2)}
\begin{figure}[htbp]
    \centering
    \includegraphics[width=1\textwidth]{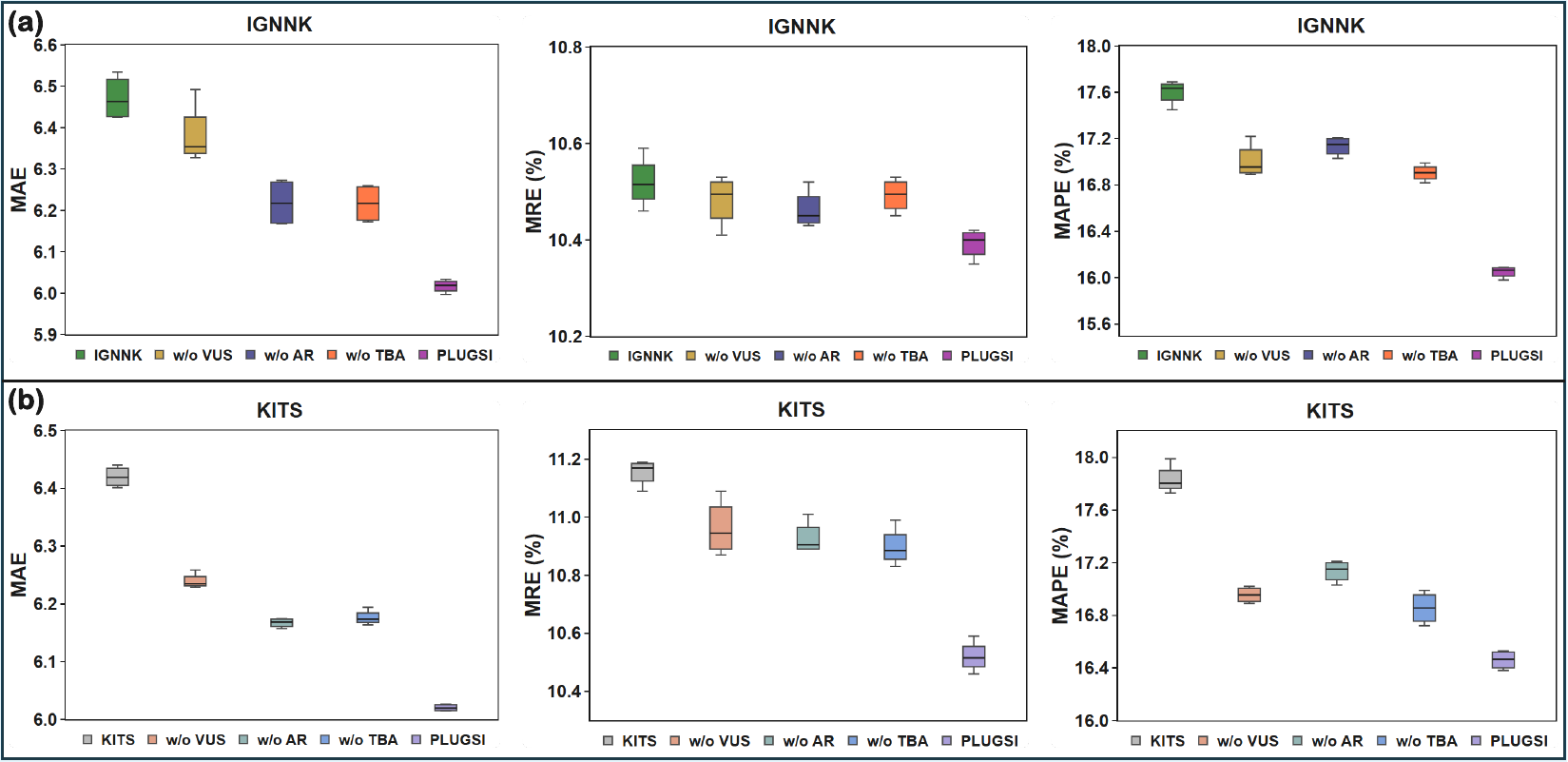}
    \caption{Ablation study for IGNNK and KITS on METR-LA dataset. As observed, each component is effective.}
    \label{fig:ablation}
\end{figure}
To answer \textbf{Q2}, we conduct ablation studies to evaluate the contribution of each key component in \textsc{PlugSI}. Specifically, we create the following variants: (1) \textbf{w/o VUS} Removes the entire VUS architecture. (2) \textbf{w/o AR} Replaces the learned adaptive adjacency matrix with a randomly initialized adaptive one. (3) \textbf{w/o TBA} Removes the entire TBA architecture. We conduct the ablation study on the METR-LA dataset using KITS and IGNNK as backbone models. Experimental results (mean ± standard deviation) are reported in Fig.~\ref{fig:ablation}. Overall, we observe that each component of \textsc{PlugSI} contributes positively to its performance. Specifically, removing the VUS results in the most significant performance drop, highlighting its crucial role in evaluating the uncertainty of virtual nodes and facilitating effective adaptation of the graph structure. Replacing the AR with a randomly initialized adaptive adjacency matrix, which lacks awareness of the local neighborhood around virtual nodes, fails to learn a well adaptation. Besides, removing TBA, which provides historical adaptation information, also leads to decreased performance, demonstrating the importance of leveraging historical information for stable test-time adaptation.

\subsection{Parameter Study (Q3)}
To answer \textbf{Q3}, we conduct parameter studies on METR-LA dataset using KITS and IGNNK. To this end, we study three key parameters: the information balance coefficient $\alpha$, the regularization weight $\lambda$, and the small-batch size $B$. Specifically, $\alpha$ determines the stability of \textsc{PlugSI}. Inspired by common strategies~\cite{tarvainen2017mean,xie2020unsupervised,sohn2020fixmatch} and considering the importance of historical adaptation information, we conduct sensitivity analysis for $\alpha$ over the set $\{0.9, 0.99, 0.995, 0.999\}$. For the regularization weights that balance the optimization objective, we evaluate $\lambda \in \{0.1, 0.5, 1, 5, 10\}$. The batch size $B$ plays an important role in adapting to streaming data and scenarios is often variable in real-world; therefore, we study $B \in \{8, 16, 32, 64, 128\}$. 
\begin{figure}[htbp]
    \centering
    \includegraphics[width=1\textwidth]{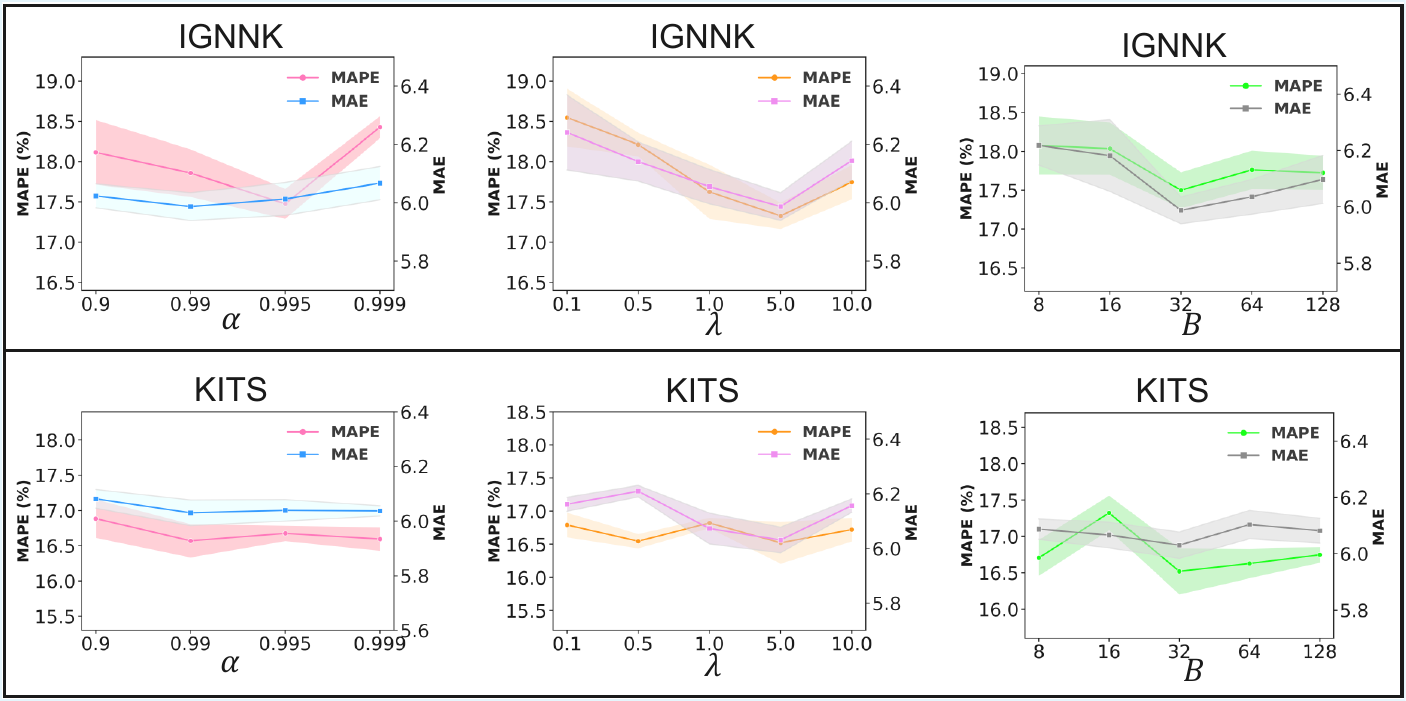}
    \caption{Impact of number of information balance coefficient $\alpha$, regularization weight $\lambda$ and small-batch size $B$ for IGNNK and KITS on METR-LA dataset.}
    \label{fig:parameter}
\end{figure}
The results are reported in Fig.~\ref{fig:parameter}. 
Overall, these results show that \textsc{PlugSI} exhibits stable behavior with respect to $\alpha$
, and achieves reliable performance under moderate $\lambda$ and $B$. Specifically, the MAE remains relatively stable as $\alpha$ increases and indicates that the performance is generally insensitive to the update rate. Because sufficient historical information is retained to ensure stable long-term memory.
For $\lambda$, both MAPE and MAE show mild fluctuations with increasing values, where moderate regularization achieves the best balance between structural adaptation and preserving the original topology.
Similarly, as the batch size $B$ increases, performance first improves and then slightly declines, suggesting that moderate batches enable more effective adaptation while overly large ones may weaken responsiveness.
\subsection{Resilience on Different Virtual Patterns (Q4)}
\begin{figure}[htbp]
    \centering
    \includegraphics[width=1\textwidth]{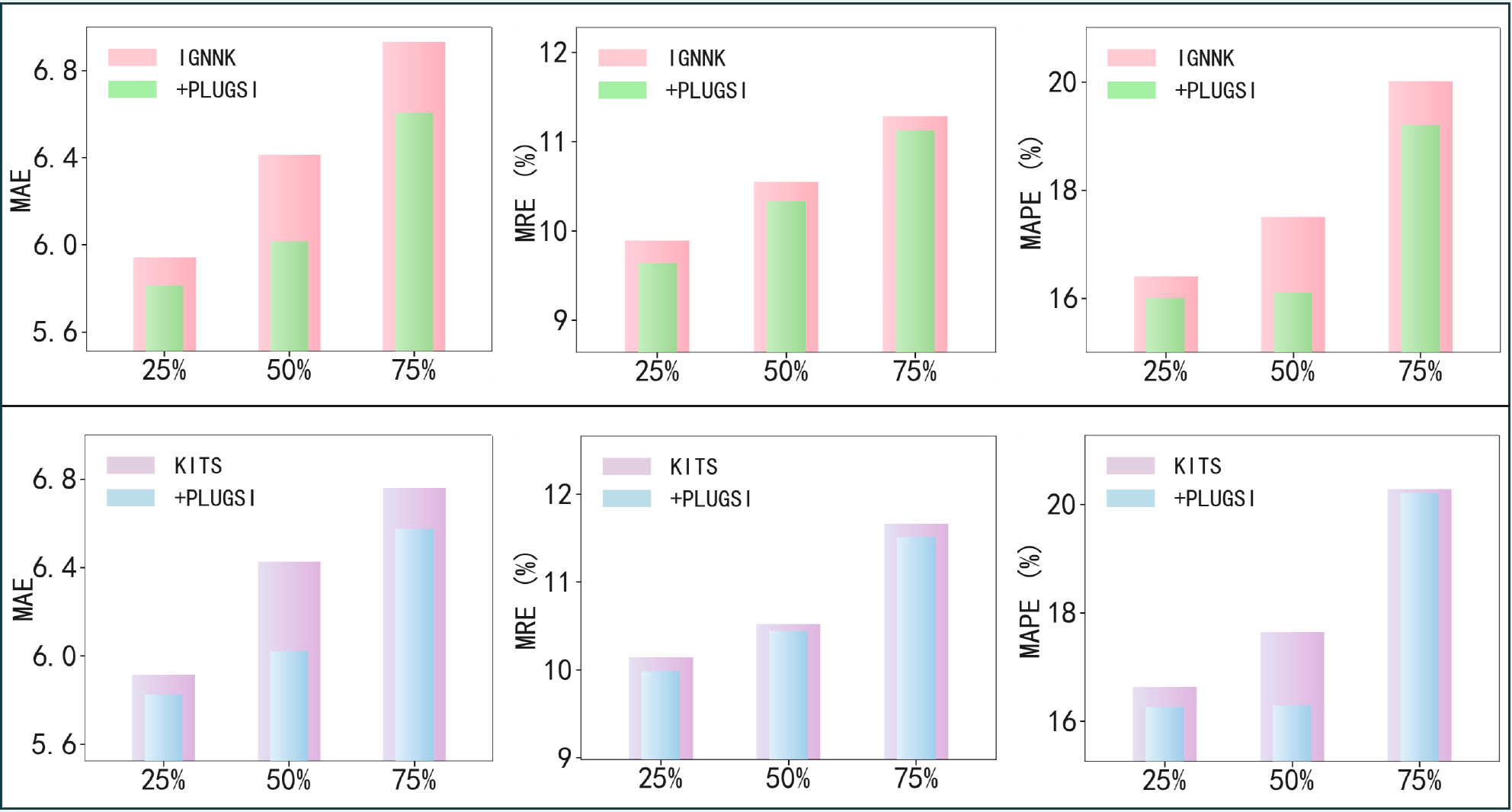}
    \caption{Experimental results under different proportions of virtual nodes for IGNNK and KITS on the METR-LA dataset.}
    \label{fig:miss}
\end{figure}

To investigate \textbf{Q4}, we conduct two types of virtual nodes experiments on the METR-LA dataset using KITS and IGNNK. 
\paragraph{\textbf{Different Proportions of Virtual Nodes.}}
In details, we conduct experiments simulating different ratios between observed and virtual nodes. By default, we adopt an observed-to-virtual ratio of 1:1 (50\%). In addition, we further evaluate the inference performance under 3:1 (25\%) and 1:3 (75\%) settings to simulate various real-world deployment proportions. Experimental results are reported in Fig. \ref{fig:miss}, revealing a consistent trend: the performance of all models
deteriorates as the virtual nodes ratio increases. However, an intriguing observation is that even with a higher proportion of virtual nodes, our model consistently refines the test graphs effectively, enabling the backbone models to achieve superior inference performance during test. This remarkable stability under challenging conditions further highlights the robustness and practical applicability of \textsc{PlugSI}.

\paragraph{\textbf{Regional Virtual-Clusters inference.}}
By default, we focus on the Random Virtual Nodes pattern, where the observed nodes from the complete dataset are randomly missing. 
In this section, we further consider a more challenging and realistic scenario, 
\begin{wraptable}{r}{0.5\textwidth} 
\setlength{\tabcolsep}{3pt}
\renewcommand{\arraystretch}{1.2}
\centering
\scriptsize
\caption{
Performance on the METR-LA dataset using IGNNK and KITS under regional virtual clusters.
}
\vspace{7pt}
\label{tb:region}
\resizebox{0.45\textwidth}{!}{%
\begin{tabular}{cccc}
\hline\hline
\textbf{Backbones} & \textbf{MAE} & \textbf{MRE} & \textbf{MAPE} \\ 
\hline
IGNNK & 10.290 & 0.189 & 0.304 \\
\rowcolor{lightroyalblue}\textsc{+ PlugSI} & \textbf{9.887} & \textbf{0.183} & \textbf{0.296} \\ 
\hline
KITS & 7.204 & 0.124 & 0.261 \\
\rowcolor{lightroyalblue}\textsc{+ PlugSI} & \textbf{6.916} & \textbf{0.120} & \textbf{0.252} \\ 
\hline\hline
\end{tabular}
}
\end{wraptable}
namely Regional Virtual-Clusters, which is also of great practical significance. 
As shown in Table~\ref{tb:region}, since most virtual nodes in this setting lack nearby observed counterparts for reference, the overall performance of backbone models declines. 
Nevertheless, after refining the test graph with \textsc{PlugSI}, the inference performance of backbones is still enhanced, highlighting that \textsc{PlugSI} remains robust and effective even under complex and adverse real-world conditions.

\subsection{Computational Cost (Q5).}
\begin{wraptable}{r}{0.58\textwidth}
\vspace{-1cm}
\renewcommand{\arraystretch}{1.2}
\centering
\caption{Computational cost on different datasets using KITS and IGNNK backbones.
All results are reported in seconds (\textbf{s}).}
\vspace{7pt}
\label{tab:time}\small
\begin{tabular}{ccccc} 
\hline\hline
\textbf{Methods} & \textbf{AQI-36} & \textbf{METR-LA} & \textbf{PEMS07} & \textbf{NREL-AL} \\ 
\hline
IGNNK & 0.002 & 0.009 & 0.010 & 0.006 \\
\textsc{+ PlugSI} & 0.081 & 0.171 & 0.362 & 0.095 \\ 
\hline
KITS & 0.009 & 0.043 & 0.054 & 0.031 \\
\textsc{+ PlugSI} & 0.071 & 0.238 & 0.178 & 0.141 \\ 
\hline\hline
\end{tabular}
\vspace{-10pt}
\end{wraptable}
To evaluate \textbf{RQ5}, we assess the computational efficiency of our proposed framework. 
\textsc{PlugSI} is implemented in a plug-and-play manner at test-time, refining the test data before the pre-trained model performs inference. We measure the test-time latency of \textsc{PlugSI} against the baseline setting, where the pre-trained model directly infers on the test data without refinement. Following the standard TTA protocol~\cite{li2023robustness}, data arrive in small-batches that require immediate adaptation; therefore, we report the average inference time per batch (in seconds). To comprehensively assess computational cost, both IGNNK and KITS are evaluated across four benchmark datasets. As shown in Table~\ref{tab:time}, even with the additional adaptation overhead introduced by \textsc{PlugSI}, the per-batch time cost remains minimal, demonstrating its high efficiency and practicality—even for deployment on resource-constrained edge devices.
\section{Limitations and Future Directions}\label{section_4}

Although our work demonstrates promising results in enhancing graph-based SI model by test-time graph adaptation, several minor limitations remain.

Our study focuses on graph-based SI scenarios, which is also the main setting adopted by most deep learning-based methods~\cite{appleby2020kriging,wu2021inductive,wu2021spatial,zheng2023increase,hu2023decoupling,li2024non,xu2025kits}. However, following prior work, they may have limitations in non-sensor scenarios that are difficult to represent as graphs, such as epidemic spread or population density estimation. In the future, we plan to explore how to model these cases.

Besides, our method follows the standard TTA protocol~\cite{li2023robustness} and is designed for streaming scenarios, where data arrives in small-batches and immediate adaptation is required. This setting reflects practical demands in real-world online systems, such as real-time traffic spatial interpolation, where waiting for more samples is often infeasible. And batching different periods may smooth the distribution between strong and weak out of distribution samples~\cite{li2023robustness}. In future work, we plan to extend this setting by incorporating accumulated batches to better utilize temporal context and improve adaptation robustness.

\section{Related Work}\label{section_5}

Spatial Interpolation (SI), also known as Spatio-temporal Kriging~\cite{oliver1990kriging}, is a prominent technique for inferring the readings of virtual sensors from observed sensor data. Early methods such as inverse distance weighting~\cite{lu2008adaptive} infer nodes based on linear dependencies. Then, nonlinear relations are captured by subsequent approaches such as Gaussian processes (GPs)~\cite{rasmussen2003gaussian}. Most previous deep learning-based SI methods~\cite{wu2021inductive,wu2021spatial,zheng2023increase,hu2023decoupling,li2024non,xu2025kits} are graph-based. They aimed to adapt the spatio-temporal patterns learned from observed nodes directly to virtual nodes. For example, INCREASE~\cite{zheng2023increase} leverages metadata such as points of interest (POIs) and functional areas to construct diverse predefined graphs and prioritizes nearest-neighbor sensors during aggregation, aiming to learn richer spatio-temporal representations that can generalize to virtual nodes in unknown test graphs. KITS~\cite{xu2025kits}, a state-of-the-art SI method, inserts virtual nodes during training in an attempt to simulate the possible topologies that may appear at test-time. However, this approach incurs high training costs and struggles to fully capture the diverse and unknown graph structures. DualSTN~\cite{hu2023decoupling} focuses on modeling long-term and short-term temporal patterns during training, but overlooks the exploitation of informative temporal dynamics available at test-time. 

These motivate us to take a different perspective: instead of simulating virtual nodes during training, we leverage the topological information introduced by virtual nodes at test-time to guide the inference of their readings. In contrast, \textsc{PlugSI} leverages the topology induced by virtual nodes for more accurate inference on unseen graphs and can be readily combined with existing graph-based models to boost their predictive performance.

\section{Conclusion}\label{section_6}
%
%
%
In this paper, we revisit SI from a test-time perspective and propose \textsc{PlugSI}, a graph data-centric framework that enhances the robustness of pre-trained SI models under streaming and evolving sensor networks. Unlike prior works that rely solely on pre-training, \textsc{PlugSI} leverages test-time graph information to refine graph structures without accessing training data or altering model parameters. Specifically, we introduce two complementary modules: the UTA, which adapts to unknown and dynamically changing graph topologies within each small batch, and the TBA, which stabilizes adaptation by balancing current updates with accumulated historical information.
Extensive experiments on four real-world datasets demonstrate that \textsc{PlugSI} consistently improves diverse graph-based SI models. Moreover, complexity analysis, runtime cost experiments, and various other robustness experiments demonstrate that \textsc{PlugSI} is robust, scalable, and lightweight, indicating high practical value.

\bibliographystyle{splncs04}
\bibliography{reference}

\appendix
\title{Appendix}
\section{Spatial Interpolation Definition}\label{section_Appdx}
Let $\mathbf{X}^o_{t:t+P} \in \mathbb{R}^{N_o \times P}$ denote the readings of $N_o$ observed sensors over $P$ time intervals.
The spatial interpolation (SI) problem aims to learn a function $\mathcal{F}$ to infer the readings
$\mathbf{X}^u_{t:t+P} \in \mathbb{R}^{N_u \times P}$ at $N_u$ virtual sensors,
which are unavailable during training and only specified at test-time,
based on the measurements $\mathbf{X}^o_{t:t+P}$ from observed sensors.
Spatial relationships among sensors are represented by an adjacency matrix
$\mathbf{A} \in \mathbb{R}^{(N_o + N_u) \times (N_o + N_u)}$,
which is constructed according to the spatial distances between sensor locations.

Under this formulation, during training, the function $\mathcal{F}$ is learned by using a subset of observed sensors
$\mathbf{X}^{o}_{t:t+P}$ to simulate virtual sensors, together with the adjacency matrix
$\mathbf{A}^{o} \in \mathbb{R}^{N_o \times N_o}$.
At test-time, $\mathcal{F}$ is applied to the observed readings
$\mathbf{X}^{o}_{t:t+P}$ and the full adjacency matrix
$\mathbf{A} \in \mathbb{R}^{(N_o + N_u) \times (N_o + N_u)}$ to infer the readings
$\hat{\mathbf{X}}^{u}_{t:t+P}$ of virtual sensors.

\end{document}